\documentclass{article}

\PassOptionsToPackage{sort, numbers, compress}{natbib}
\usepackage[preprint]{neurips_2025}

\usepackage[utf8]{inputenc} % allow utf-8 input
\usepackage[T1]{fontenc}    % use 8-bit T1 fonts
\usepackage{hyperref}       % hyperlinks
\usepackage{url}            % simple URL typesetting
\usepackage{booktabs}       % professional-quality tables
\usepackage{amsfonts}       % blackboard math symbols
\usepackage{nicefrac}       % compact symbols for 1/2, etc.
\usepackage{microtype}      % microtypography
\usepackage{multicol}
\usepackage{multirow}
\usepackage{makecell}
\usepackage{enumitem}
\usepackage[table]{xcolor}
\usepackage{graphicx}
\usepackage{pifont}
\usepackage{wrapfig}
\usepackage{amsmath, amssymb}
\usepackage[most]{tcolorbox}
\usepackage{geometry}
\usepackage{booktabs}
\usepackage{subcaption}
\usepackage{fontawesome}

\definecolor{softgreen}{RGB}{34,139,34}

\title{GThinker: Towards General Multimodal Reasoning via Cue-Guided Rethinking}

\author{%
Yufei Zhan\textsuperscript{1,3,}$^{\dagger,}$\thanks{Work done during internship at ByteDance.}, Ziheng Wu\textsuperscript{2,}\thanks{Equal Contribution.}\;\;, Yousong Zhu\textsuperscript{1,\faEnvelopeO}, \\
\textbf{Rongkun Xue}\textsuperscript{6}, \textbf{Ruipu Luo}\textsuperscript{2}, \textbf{Zhenghao Chen}\textsuperscript{2}, \textbf{Can Zhang}\textsuperscript{2}, \textbf{Yifan Li}\textsuperscript{7}, \textbf{Zhentao He}\textsuperscript{2}, \\\textbf{Zheming Yang}\textsuperscript{2}, \textbf{Ming Tang}\textsuperscript{1,3}, \textbf{Minghui Qiu}\textsuperscript{2}, \textbf{Jinqiao Wang}\textsuperscript{1,3,4,5}\\
{$^{1}$ Foundation Model Research Center, Institute of Automation, Chinese Academy of Sciences}\\
{$^2$ ByteDance}\;\;
{$^{3}$ School of Artificial Intelligence, University of Chinese Academy of Sciences}\\
{$^4$ Peng Cheng Laboratory}\;\;{$^5$ Wuhan AI Research}\\{$^6$Xi’an Jiaotong University}\;\;{$^7$Renmin University of China}\\
{\tt \url{https://github.com/jefferyZhan/GThinker}}
}

\begin{document}

\maketitle

\begin{abstract}

    Despite notable advancements in multimodal reasoning, leading Multimodal Large Language Models (MLLMs) still underperform on vision-centric multimodal reasoning tasks in general scenarios. This shortfall stems from their predominant reliance on logic- and knowledge-based ``slow thinking'' strategies—while effective for domains like math and science—fail to integrate visual information effectively during reasoning. Consequently, these models often fail to adequately ground visual cues, resulting in suboptimal performance in tasks that require multiple plausible visual interpretations and inferences. To address this, we present \textbf{GThinker} (General Thinker), a novel reasoning MLLM excelling in multimodal reasoning across general scenarios, mathematics, and science. GThinker introduces Cue-Rethinking, a flexible reasoning pattern that grounds inferences in visual cues and iteratively reinterprets these cues to resolve inconsistencies. Building on this pattern, we further propose a two-stage training pipeline, including pattern-guided cold start and incentive reinforcement learning, designed to enable multimodal reasoning capabilities across domains. Furthermore, to support the training, we construct \textbf{GThinker-11K}, comprising 7K high-quality, iteratively-annotated reasoning paths and 4K curated reinforcement learning samples, filling the data gap toward general multimodal reasoning. Extensive experiments demonstrate that GThinker achieves 81.5\% on the challenging comprehensive multimodal reasoning benchmark M$^3$CoT, surpassing the latest O4-mini model. It also shows an average improvement of 2.1\% on general scenario multimodal reasoning benchmarks, while maintaining on-par performance in mathematical reasoning compared to counterpart advanced reasoning models. The code, model, and data will be released soon at \url{https://github.com/jefferyZhan/GThinker}.
\end{abstract}
\begingroup
\renewcommand*{\addcontentsline}[3]{}
\section{Introduction}
\label{intro}

\begin{figure}
  \centering
   %\fbox{\rule{0pt}{2in} \rule{0.9\linewidth}{0pt}}
   % \vskip 0.2in
   \includegraphics[width=\linewidth]{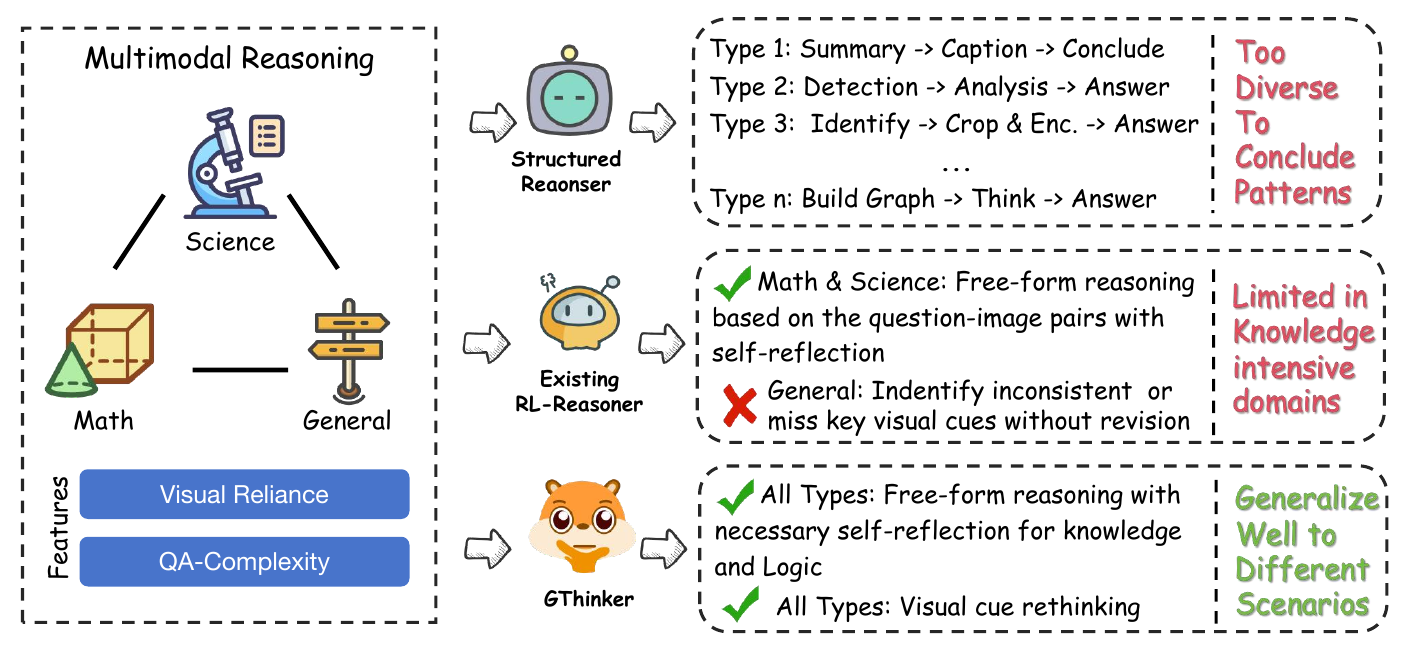}
   \caption{Multimodal reasoning methods comparison across scenarios. Multimodal Reasoning in different domains is featured with visual reliance and high question complexity, making it a challenging task. Different from previous methods, GThinker utilizes free-form thinking for different types of questions instead of a fixed structure form and enables general scenario reasoning accuracy with designed visual cue rethinking.}

   \label{fig:page}
   % \vskip -0.2in
\end{figure}

Open-source Multimodal Large Language Models (MLLMs) \cite{llavaonevision, wu2024deepseekvl, zhu2025internvl3,wu2025valley2,team2025kimi} have made significant strides across a wide range of tasks. Leading models like Qwen2.5-VL \cite{Qwen2.5-VL} now rival closed-source counterparts such as GPT-4o \cite{hurst2024gpt4o} in performance. These advances have benefited in part from the adoption of chain-of-thought (CoT) techniques \cite{lu2022learn, yao2023tree, wei2022chain}, especially in mathematics and science. With the emergence of OpenAI’s O1 model \cite{jaech2024openai}, several studies \cite{yao2024mulberry, xu2024llavacot, thawakar2025llamav} have sought to transfer such slow-thinking capabilities to the multimodal reasoning domain to enhance models’ performance on complex tasks. DeepSeek-R1 \cite{guo2025deepseekr1} further introduces a new perspective, showing that outcome-reward Reinforcement Learning (RL) can awake long CoT reasoning, with promising results \cite{meng2025mmeureka, yang2025r1one, chen2025VLAA} in multimodal reasoning tasks involving science and mathematics.

Beyond mathematics and science, multimodal reasoning in general scenarios, which often involves visual cues and related commonsense still remains under-explored. Unlike math and science tasks, which typically follow strict logical structures and have unique answers, multimodal reasoning tasks in general scenarios are more diverse in nature. This makes it challenging to summarize a fixed CoT pattern or design an effective Process Reward Model (PRM), limiting the effectiveness of structured reasoning \cite{yao2024mulberry, xu2024llavacot} and Multimodal PRMs \cite{wang2025visualprm, liu2024diving}. Furthermore, general scenarios often require plausible interpretations and inferences grounded in visual content, which reduces the effectiveness of current outcome-reward-based reasoning models \cite{huang2025vision, yang2025r1one, chen2025VLAA} that are primarily developed for math and science. As summarized in Figure \ref{fig:page}, existing slow-thinking models frequently miss critical visual cues. When encountering plausible but inconsistent outputs, they often proceed directly to an answer without revisiting the reasoning path, unlike the reflection and verification behaviors observed in math and science domains. This suggests that in general scenarios, rethinking that integrates visual interpretation and inference cannot be effectively incentivized by RL alone, in contrast to the naturally learned reflection mechanisms in math and science reasoning tasks during pretraining \cite{shah2025rethinking}.

To address these challenges, we propose GThinker, a novel reasoning MLLM excelling in multimodal reasoning across general scenarios, mathematics, and science. First, we introduce a new long-chain cue-driven pattern for multimodal reasoning called Cue-Rethinking. Unlike prior approaches \cite{xu2024llavacot, thawakar2025llamav} that define structured CoT formats, Cue-Rethinking only requires the reasoning process to be strictly grounded in visual cues without enforcing a fixed format. After completing an initial reasoning chain, the model rethinks on the interpretations and inferences based on visual content to correct inconsistencies and arrive at the correct answer. Building on this pattern, we propose a two-stage training pipeline to enable robust multimodal reasoning. We begin by using pattern-guided cold start to train the model to learn this reasoning pattern on different tasks, and cold-start it with supervised fine-tuning. Then, we further employ an incentive RL stage to let the model explore optimal strategies for solving diverse problems across domains. To support training, we further develop a multimodal iterative annotation pipeline based on the latest advancing multimodal models like O3 \cite{openai2025o3o4mini} and construct GThinker-11k, compromised 7K cold-start data with high-quality annotated reasoning paths and 4K reinforcement learning samples, filling a key gap in multimodal reasoning fine-grained data for general scenarios.

We implement GThinker based on the advanced open-source MLLM Qwen-VL 2.5–7B and conduct extensive experiments to rigorously evaluate its effectiveness. We first benchmark GThinker against both open- and closed-source models on M$^3$CoT \cite{chen2024m3}, a challenging and comprehensive multimodal reasoning dataset spanning science, general commonsense, and mathematics. For broader validation, we include general-domain benchmarks such as MMStar \cite{chenmmstar} and RealWorld QA \cite{xai2024realworldQA}, as well as science and math-focused benchmarks including MMMU-Pro \cite{yue2024mmmu-pro}, MathVision \cite{wang2024mathvision}, and MathVista \cite{lumathvista}. GThinker demonstrates strong performance across all domains, achieving 81.5\% on M$^3$CoT—surpassing the advanced O4-mini model. On MMStar and RealWorld QA, GThinker achieves the improvement of 2.5\% and 1.6\%, respectively. Additionally, it performs competitively on science and math benchmarks with 40.7\% on MMMU Pro and 72.7\% on MathVista, matching or outperforming recent RL-enhanced approaches, further validating its effectiveness.

\section{Related Work}
\subsection{Structured Multimodal Chain-of-Thought Reasoning}

Structured Multimodal Chain-of-Thought (MCoT) reasoning builds on the Chain-of-Thought (CoT) paradigm \cite{wei2022chain}, extending it to multimodal tasks using step-by-step reasoning \cite{lu2022learn, zhang2023multimodalcot}. Many approaches enhance this framework with structured designs \cite{zheng2023ddcot, liu2024chain, mitra2024compositional} and further improvements such as fine-grained visual grounding, context integration, or tool use \cite{jia2024dcot, gao2024cantor, luan2024textcot, wu2024vstar, shao2024visual, li2024vocot, bigverdi2024perception}. However, these methods are often task-specific—e.g., CCoT \cite{mitra2024compositional} for compositional reasoning, LLaVA-Aurora \cite{bigverdi2024perception} for spatial reasoning—and lack robustness across diverse scenarios. Recently, slow-thinking paradigms \cite{jaech2024openai, qwq32b, qin2024o1journey} have been proposed to improve reasoning depth. Enhanced MCoT variants like LLaVA-CoT \cite{xu2024llavacot}, Virgo \cite{du2025virgo}, and Mulberry \cite{yao2024mulberry} leverage long-chain generation, tree search, and self-reflection. Yet, they remain confined to structured, logic-heavy tasks and are difficult to generalize to broader settings. In contrast, GThinker adopts a 
free-form, cue-based thinking paradigm with further visual cue-based rethinking, moving beyond rigid structures to support open-domain multimodal reasoning. This design enables generalization across task types without sacrificing interpretability or performance.

\subsection{Multimodal Reasoning with Reinforcement Learning}

Reinforcement learning (RL) has become a powerful tool to align MLLMs and mitigate hallucinations \cite{sun2023aligning, yu2024rlaif, zhang2024spa, fact-rlhf, li2023silkie, zhang2025mm-rlhf}, and is now being explored to improve multimodal reasoning. Early approaches like LLaVA-Reasoner \cite{zhang2024improve} and MPO \cite{wang2024enhancing} rely on rationale distillation alone and preference data to guide reasoning, while Insight-V \cite{rafailov2023direct} designs multi-agent systems with iterative Direct Preference Optimization. However, these methods focus on ``teaching correctness'' via supervised signals and human preference annotations, limiting robustness and scalability for more complex scenarios. A shift emerged with DeepSeek-R1 \cite{guo2025deepseekr1}, which showed that outcome-based rewards, without fine-grained annotations, can drive reasoning through self-verification and reflection. Follow-up works \cite{chen2025r1v, yang2025r1one, meng2025mmeureka, chen2025VLAA, team2025kimi, peng2025skywork} expand this idea to the multimodal domain, leveraging verifiable reward functions or rule-based signals to improve math and science reasoning. Yet, these methods largely target well-defined tasks with unique answers. In general multimodal reasoning, models must handle ambiguity, interpret visual cues, and perform flexible inference. This limits the direct transfer of knowledge-style RL setups. Additionally, common reward models like PRMs \cite{wang2025visualprm, liu2024diving} struggle to capture progress in diverse tasks under general scenarios. To address this, we propose a clue-driven rethinking pattern tailored for general scenario multimodal scenarios but also accustomed to math and science settings. By further leveraging our design two-stage training, GThinker enables flexible reasoning with visual cue-based rethinking and knowledge reflection across diverse multimodal reasoning tasks.

\section{Methodology}
In this section, we provide a comprehensive description of the novel multimodal reasoning model GThinker as depicted in Figure \ref{fig:GThinker}. In \S\ref{sec: pattern}, we first present the Cue-Rethinking Pattern, a core component built on free-form thinking to provide visual cue-driven guidance for multimodal reasoning across scenarios. Then, in \S\ref{sec: learning}, we describe Pattern-Guided Cold Start, in which we build 7k high-quality reasoning path annotated data and train the model with pattern-guided supervised fine-tuning to learn how to think and rethink for different scenarios. Finally, we introduce Incentive Reinforcement Learning to generalize the multimodal reasoning capabilities of the model across diverse scenarios in \S\ref{sec: rl}.

\begin{figure}
  \centering
   %\fbox{\rule{0pt}{2in} \rule{0.9\linewidth}{0pt}}
   % \vskip 0.2in
   \includegraphics[width=\linewidth]{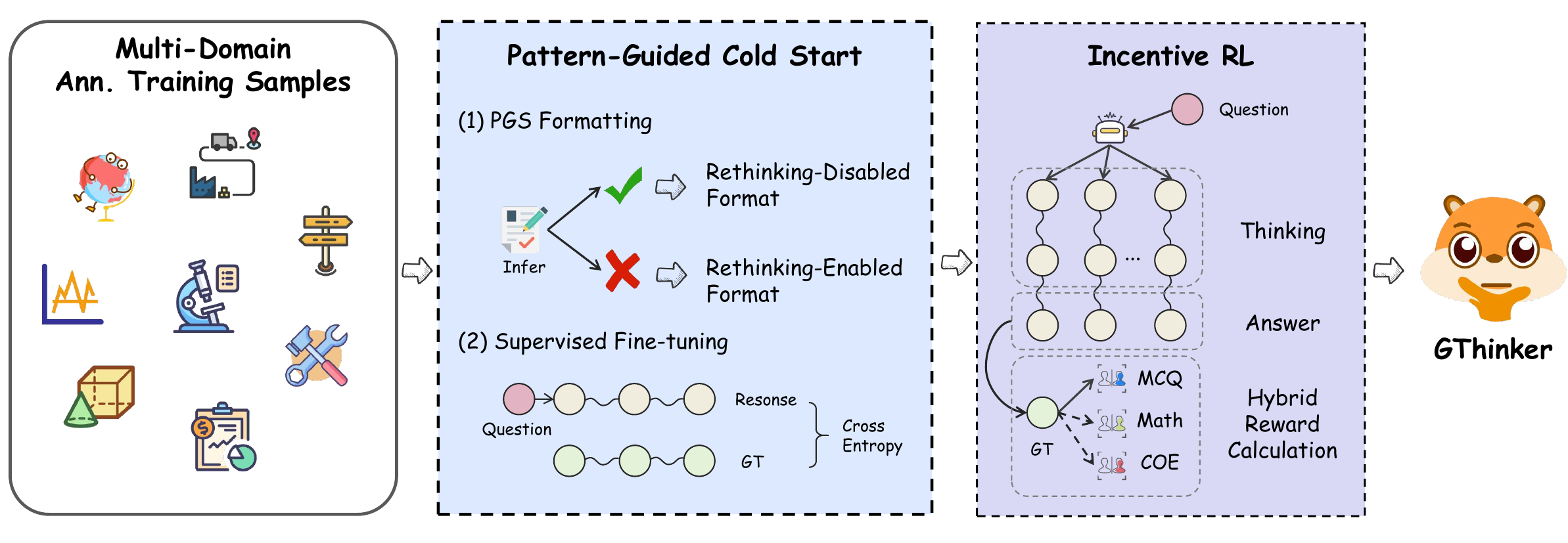}
   \caption{{Overall pipeline for constructing GThinker.} We collect multi-domain data covering general, math, and science tasks, and annotate it using multiple advanced MLLMs. The Pattern-Guided Cold Start phase then teaches the model the Cue-Rethinking Pattern for different question types. Finally, incentive reinforcement learning with DAPO enhances GThinker’s ability to perform adaptive and accurate multimodal reasoning across diverse scenarios.}
   \label{fig:GThinker}
   % \vskip -0.2in
\end{figure}

\subsection{Cue-Rethinking Pattern}
\label{sec: pattern}

Existing long-chain reasoning methods\cite{xu2024llavacot, thawakar2025llamav} often rely on fixed, structured thinking chains tailored to specific tasks. While effective in targeted domains, their performance tends to drop sharply when applied to more general or unfamiliar scenarios. Outcome-reward models offer more flexibility, but they also fall short in general settings that require grounded, visually informed interpretations and inferences. To tackle this challenge, we introduce the Cue-Rethinking Pattern, a thinking framework that enables flexible long-chain reasoning through a combination of free-form thinking and rethinking on visual cues.

\begin{wrapfigure}{r}{0.52\textwidth}  % r 表示右侧，0.4 表示占页面宽度的 40%
  \centering

  \includegraphics[width=0.52\textwidth]{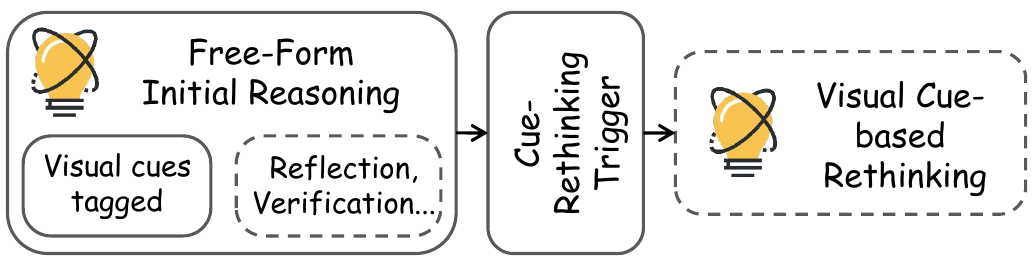} % 使用你自己的图片路径
  \caption{Toy example of the Cue-Rethinking Pattern. The dashed line indicates generation on demand.}
  \label{fig: pattern}
\end{wrapfigure}

As shown in Figure \ref{fig: pattern}, this process unfolds in three stages generally: initial reasoning, cue-rethinking trigger, and cue-based rethinking. During the initial stage, the model is free to reason in any form based on the question and image content itself, without structural constraints. It simply tags any referenced visual cues in the format <vcues\_*> </vcues\_*> (* indicates the No.), which are later used for visual cues rethinking. This flexibility allows the model to apply learned reasoning strategies, such as step-by-step deduction or logical and knowledge reflection, much like how reasoning is approached in mathematical or scientific contexts, depending on the task.

After completing the initial reasoning, a prompt is triggered to initiate cue-based rethinking, like \textit{``Let's check each visual cue and corresponding reasoning before reaching the final answer''}. Importantly, we do not require immediate rethinking after visual cue identification, as doing so may disrupt the natural reasoning flow and prevent us from seeing the overall context. Then, the model revisits all previously marked visual cues, checking for inconsistencies or flaws. If problematic cues are identified, they are revised, and the model re-engages in corresponding reasoning, now grounded in the corrected cues, and concludes the final answer. This approach not only accommodates a wide range of reasoning approaches for different tasks but also addresses current limitations in handling misleading or missing visual inputs during reasoning. By combining free-form thinking with designed visual cue rethinking, this pattern delivers robust, adaptable reasoning across diverse multimodal reasoning scenarios.

\subsection{Pattern-Guided Cold Start}
\label{sec: learning}

Building on the Cue-Rethinking Pattern, we address how to effectively teach models to internalize and apply this reasoning pattern. While outcome-reward RL can guide models toward desired thinking, relying solely on it is still challenging and computationally intensive \cite{guo2025deepseekr1}. To overcome this, we introduce a Pattern-Guided Cold Start stage as shown in Figure \ref{fig:GThinker}, where the model is trained to adopt the Cue-Rethinking paradigm through supervised fine-tuning. To support this, we construct a 7K-scale dataset of annotated reasoning paths across multiple domains, using a multimodal collaborative annotation pipeline. The resulting data enables the model to learn both general problem-solving and cue-based rethinking.

\begin{figure}
  \centering
   %\fbox{\rule{0pt}{2in} \rule{0.9\linewidth}{0pt}}
   % \vskip 0.2in
   \includegraphics[width=0.85\linewidth]{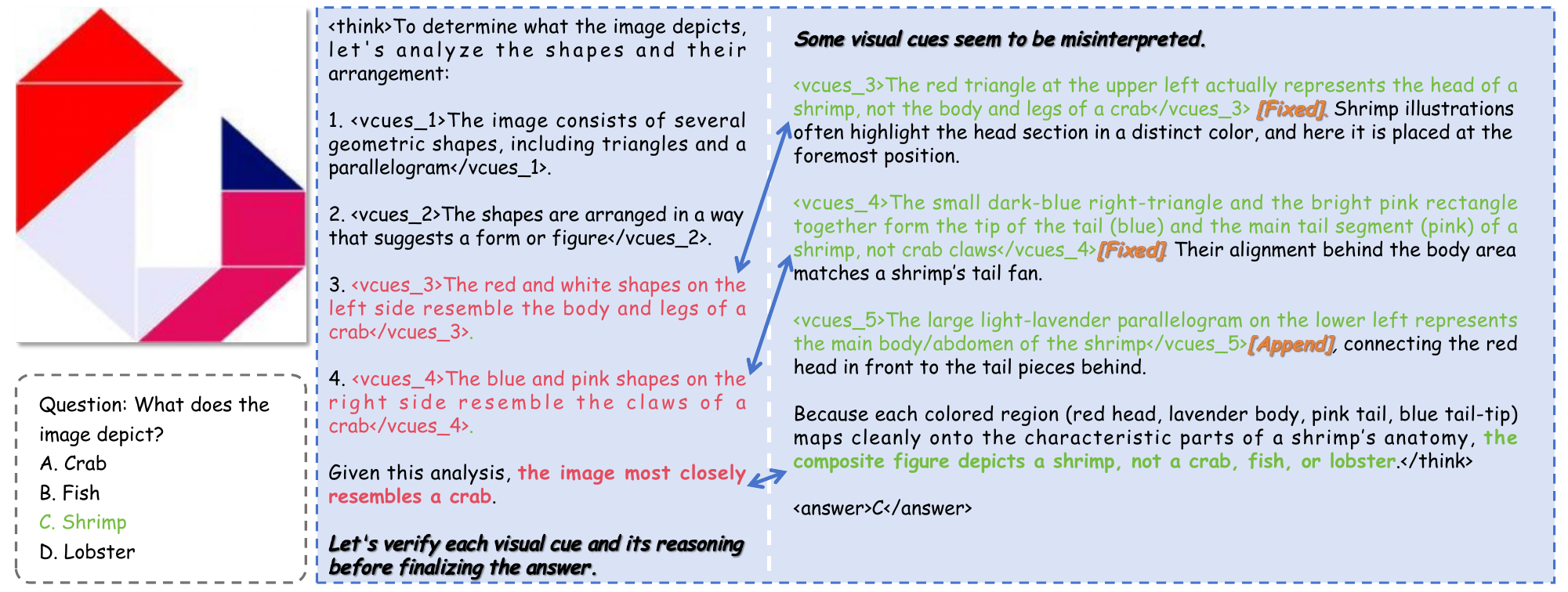}
   \caption{{Constructed Data Example with Cue Rethinking.} The visual cues in red are flawed ones, while the green indicates the visual cues are revised or appended.}
   \label{fig:data}
   % \vskip -0.2in
\end{figure}

\textbf{Data Construction via Multimodal Iterative Annotation.} To support domain-diverse multimodal reasoning, we collect data spanning mathematics, science, and general visual scenarios, validating each example for visual dependency and reasoning complexity. Instead of prompting models with image captions and question or structure requirements, we feed the image, question, and answer into advanced multimodal reasoning models, prompting them to reason step-by-step and identify relevant visual cues. For math and science questions, the models are allowed with self-reflection and validation; for cue-rich general questions, they are instructed to provide explicit visual references to support later rethinking. This strategy aligns well with the flexible design of Cue-Rethinking. To maximize precision, we iteratively annotate the data using several models, including GPT-4o, o1, and o3, leveraging each model’s strengths. We further extend this process to generate cue-based rethinking data. This automated pipeline results in a final dataset of 7,358 high-quality annotated samples, detailed further in the Appendix A. We provide a data example with key texts formatted in cue-rethinking in Figure \ref{fig:data}.

\textbf{Pattern-Guided Supervised Fine-tuning.} With the annotated data, we train the model to learn the Cue-Rethinking pattern via supervised fine-tuning. Since reflection in science and math scenarios or cue-rethinking in general scenarios is one of the reasoning approaches, enforcing a single learning format could constrain the model's robustness. To address this, we introduce \textbf{pattern-guided selective formatting} to customize the training data based on problem type. Specifically, we first run the base model on the training questions and compare its reasoning paths to the annotations. Samples with flawed visual cues are selected to form full Cue-Rethinking sequences, covering all three stages. Remaining examples are formatted as free-form reasoning paths. The model is then fine-tuned using this pattern-compiled data, enabling it to adaptively perform reasoning or rethinking as required by the question.

\subsection{Incentive Reinforcement Learning} 
\label{sec: rl}

Following the Pattern-Guided Cold Start phase, the model acquires the designed reasoning pattern and learns to perform both flexible step-by-step reasoning and cue-based rethinking. Building on this foundation, we further enhance the model using outcome-reward reinforcement learning to encourage exploration and help it generalize across diverse tasks and scenarios. Given recent advances in outcome-reward reinforcement learning, we adopt the Decoupled Clip and Dynamic Sampling Policy Optimization (DAPO) algorithm \cite{yu2025dapo} due to its strengths in supporting long-chain reasoning and its efficiency in stable training. To accommodate varying task types and align with our pattern-based methodology, we design a hybrid reward computation strategy tailored to different problem categories. This training is carried out on a curated set of 4K diverse reasoning samples, enabling the model to generalize beyond the supervised data and adapt effectively to new challenges. %[This approach allows the model to dynamically select the most effective reasoning strategy for each case, improving its robustness and adaptability.]

\textbf{Preliminaries about DAPO.}
DAPO improves from the Group Relative Policy Optimization (GRPO) \cite{shao2024deepseekmath} with several enhancements to improve training efficiency, stability, and long-chain benefits, while retaining the key features such as outcome-based reward and policy optimization. As shown in the Equation \ref{eq: dapo}, DAPO first employs a clip-higher strategy to address exploration limitations caused by identical responses, by adjusting the clipping threshold. It then adopts a dynamic sampling mechanism to prevent low training efficiency when all responses in a group are either entirely correct or entirely incorrect. Furthermore, it integrates Token-Level Policy Gradient Loss to encourage the model to learn high-quality reasoning patterns within long-chain responses while suppressing redundant reasoning. Lastly, the Overlong Reward Shaping strategy helps reduce the noise caused by excessively long sample sequences during training.
\begin{equation}
\begin{aligned}
\mathcal{J}_{\text{DAPO}}(\theta) &= 
\mathbb{E}_{(q,a)\sim\mathcal{D}, \{o_i\}_{i=1}^{G} \sim \pi_{\text{old}}(\cdot|q)}\\
&\left[
\frac{1}{\sum_{i=1}^{G} |o_i|}
\sum_{i=1}^{G} \sum_{t=1}^{|o_i|} 
\min\left( r_{i,t}(\theta) \hat{A}_{i,t},\,
\text{clip}\left(r_{i,t}(\theta), 1 - \varepsilon_{\text{low}}, 1 + \varepsilon_{\text{high}} \right) \hat{A}_{i,t} \right)
\right]\\
\text{s.t.} \quad& 0 < \left| \left\{ o_i \mid \texttt{is\_equivalent}(a, o_i) \right\} \right| < G,
\end{aligned}
\label{eq: dapo}
\end{equation}

\noindent
{where}
\begin{align}
r_{i,t}(\theta) &= 
\frac{\pi_{\theta}(o_{i,t} \mid q, o_{i,<t})}
     {\pi_{\theta_{\text{old}}}(o_{i,t} \mid q, o_{i,<t})}, 
\hat{A}_{i,t} = 
\frac{R_i - \text{mean}(\{R_i\}_{i=1}^{G})}
     {\text{std}(\{R_i\}_{i=1}^{G})}.
\end{align}

By incorporating DAPO, especially its clip-higher mechanism and token-level loss, the model is better equipped to sample diverse reasoning paths. This enables it to learn reasoning strategies such as reflective knowledge inference for math tasks or cue-based rethinking in general multimodal scenarios. As a result, the model improves the ability to dynamically select the most suitable reasoning strategy for each situation, improving both generalization and robustness across domains.

\textbf{Hybrid Reward Design.} The default DAPO setting combines format-based and accuracy-based rewards. Prior approaches often constrain QA tasks to rigid formats, such as multiple-choice, and depend on exact string matching to assess correctness. This limits the range of question types the model can handle, especially in general scenarios, and model-based verification further reduces training efficiency. To overcome these limitations, we propose a hybrid reward strategy within the constraints of verifiable rewards. We support three main question types: multiple-choice, math, and simple open-ended formats. For multiple-choice questions, we apply exact answer matching. For math problems—whether numeric or symbolic—we use Math-Verify \cite{mathverify} to extract and verify answers. For open-ended questions that yield concise responses (e.g., a word or short phrase), we guide the model to summarize the answer in a standardized, concise format, enabling straightforward matching during reward computation. This design expands the diversity of supported question types while preserving reward accuracy. For the format reward, we follow prior work by enforcing and verifying adherence to the think-answer structure.

\textbf{Data Construction.} To support the reinforcement learning stage, we construct a set data of 4k samples spanning math, science, and general reasoning tasks. Rather than relying solely on the 7k examples from the cold start phase, we introduce 4k samples sourced from public datasets to enhance diversity and generalization. This combined dataset offers a well-balanced and domain-spanning resource tailored for incentive RL. We provide more details about this data in the Appendix A.

\section{Experiments}
\subsection{Implementation Details}
\label{sec:imple}
\textbf{Training Settings.} We implement GThinker with the advanced MLLM Qwen2.5-VL-7B \cite{Qwen2.5-VL}, one of the latest and most capable models at this scale, combining strong visual understanding with broad general knowledge. We train the GThinker using our design two-stage pipeline, including pattern-guided cold start and incentive reinforcement learning with the constructed data. For Pattern-Guided Cold Start, we use a global batch size of 128 and a learning rate of 5e-6, training the model with the 7K reasoning path annotated data for 3 epochs. In the Incentive RL stage, we set the rollout number to 16, use a global batch size of 64, and start with a learning rate of 1e-6, training for 170 steps using the curated 4K data. Training is conducted on 4 nodes, each with 8 NVIDIA H100 GPUs. The total training time is about 9 hours. We provide more details in Appendix B.

\textbf{Evaluation Settings.} We evaluate our model against top closed-source models, including the latest O4-mini, as well as open-source base and reasoning models with comparable parameter sizes trained using diverse methodologies. The evaluation focuses on multimodal reasoning across general, mathematical, and scientific scenarios:
\begin{itemize}[left=1em]
\item \textbf{M$^3$CoT:} A challenging benchmark that spans science, commonsense, and math domains, with each example verified to require multi-step reasoning. We primarily use this benchmark to comprehensively evaluate models’ multimodal reasoning capabilities across diverse scenarios.
\item \textbf{General scenario benchmarks:} MMStar \cite{chenmmstar} and RealWorld QA \cite{xai2024realworldQA}. These benchmarks focus on general and realistic scenarios, including parts of understanding-based reasoning tasks, and are used to evaluate multimodal reasoning capabilities.
\item \textbf{Science and math scenario benchmarks:} We use MMMU-Pro \cite{yue2024mmmu-pro}, which covers multiple scientific subjects, to evaluate multimodal reasoning in scientific contexts. For math-specific evaluation, we adopt the widely used MathVista \cite{lumathvista} and MathVision \cite{wang2024mathvision}benchmarks.
\end{itemize}
All evaluations are conducted on a single node equipped with 8 NVIDIA H100 GPUs. For M$^3$CoT, we follow each model's official settings and prompts and use VLMEvalKit \cite{duan2024vlmevalkit} for fair evaluation. For other benchmarks, we use the results reported in their original papers. For RL-enhanced reasoning models, which primarily focus on math and science domains, we follow their released models and evaluation guidelines to conduct testing.
  
\begin{table}[t]
    \caption{Main results on comprehensive multimodal reasoning benchmark M$^3$CoT. Abbreviations used in the table: Lang. (Language), Nat. (Natural), Soc. (Social), Phys. (Physical), Temp. (Temporal), Alg. (Algebra), Geom. (Geometry), Theo. (Theory). Excluding closed-source models, values in bold represent the highest performance, while underlined values indicate the second-best performance across all models.}
    \centering
    \label{tab: M$^3$CoT}
    \fontsize{9pt}{\baselineskip}\selectfont
    \setlength{\tabcolsep}{4.2pt}
\begin{tabular}{lcccccccccc}
\toprule
\multirow{2}{*}{Model} & \multicolumn{3}{c}{Science} & \multicolumn{3}{c}{Commonsense} & \multicolumn{3}{c}{Mathematics} & \multirow{2}{*}{Overall}\\
\cmidrule(lr){2-4}\cmidrule(lr){5-7}\cmidrule(lr){8-10}
& {Lang.} & {Nat.} & {Soc.} & {Phys.} & Soc. & Temp. & Alg. & Geom. & Theo. & \\
\midrule
\multicolumn{11}{l}{\textbf{\textit{Closed-Source Models}}} \\
\rowcolor{gray!20} Gemini-2.5 Pro \cite{google2025gemini25pro} & 97.6 & 91.6 & 75.3 & 92.2 & 81.4 & 94.3 & 81.1 & 78.8 & 61.9 & 85.9 \\
\rowcolor{gray!20} O3-20250416 \cite{openai2025o3o4mini} & 96.2 & 89.3 & 68.0 & 91.1 & 80.2 & 93.5 & 95.0 & 87.5 & 90.5 & 83.8 \\
\rowcolor{gray!20} O4-mini-20250416 \cite{openai2025o3o4mini} & 97.2 & 84.7 & 62.9 & 94.4 & 82.6 & 91.1 & 92.9 & 86.3 & 76.2 & 80.9\\
\rowcolor{gray!20} GPT-4o-20241120 \cite{hurst2024gpt4o}& 96.7 & 72.0 & 58.3 & 91.1 & 76.4 & 82.9 & 21.4 & 31.3 & 23.8 & 67.4\\
\midrule
\multicolumn{11}{l}{\textbf{\textit{Open-Source Models}}} \\
InternVL-2.5-8B \cite{chen2024internvl}& 82.5 & 63.7 & 45.2 & 86.7 & 79.8 & 93.4 & 42.8 & 27.5 & 33.3 & 61.8\\
Ovis2-8B \cite{lu2024ovis} & 80.6 & 63.1 & 46.2 & 83.3 & 79.3 & 87.8 & 45.0 & 42.5 & 38.9 & 61.9 \\
Valley2\cite{wu2025valley2} & 85.3 & 64.4 & 48.4 & \underline{90.0} & 77.7 & 80.5 & 43.6 & 36.3 & 47.6 & 62.8\\
Qwen2.5-VL-7B \cite{Qwen2.5-VL}&82.9 & 61.2 & 46.8 & 82.2 & \underline{81.4} & 81.3 & \underline{57.9} & 40.0 & \underline{61.9} & 62.4\\
\midrule
\multicolumn{11}{l}{\textbf{\textit{Reasoning Models}}} \\
LLaVA-CoT-11B \cite{xu2024llavacot}& 72.0 & 56.4 & 41.7 & 84.4 & 72.3 & 82.1 & 37.9 & 36.3 & 33.3 & 56.0 \\
InternVL2.5-MPO-8B \cite{wang2024mpo}& \underline{92.4} & \underline{75.9} & \underline{61.9} & 85.6 & 82.6 & \underline{94.3} & 55.0 & \underline{43.8} & \underline{76.2} & \underline{73.3}\\
Kimi-VL-A3B-Thinking \cite{team2025kimi} & 86.2 & 64.4 & 39.6 & \textbf{91.1} & 78.9 & 89.4 & 13.5 & 15.0 & 14.2 & 58.3 \\
MM-Eureka-7B \cite{meng2025mmeureka}& 86.7 & 71.5 & 57.3 & 81.1 & 80.2 & 90.2 & 40.0 & 23.8 & 28.6 & 67.4\\
R1-OneVision-7B \cite{yang2025r1one} & 74.9 & 66.4 & 51.4 & 84.4 & 72.3 & 85.4 & 30.0 & 31.3 & 42.9 & 61.8\\
VLAA-Thinker-7B \cite{chen2025VLAA} & 91.0 & 70.6 & 58.1 & 78.9 & 78.1 & 87.8 & 45.7 & 35.3 & 28.6 & 68.0\\
\midrule
GThinker-7B & \textbf{92.4} & \textbf{90.7} & \textbf{68.9} & 82.2 & \textbf{81.4} & \textbf{94.3} & \textbf{73.5} & \textbf{62.5} & \textbf{81.0} & \textbf{81.5}\\
\bottomrule
\end{tabular}
\end{table}

\begin{table}[]
    \centering
    \caption{Main results on math-related and multidisciplinary benchmarks, and also fine-grained understanding of multimodal benchmarks incorporating reasoning. We use the setting detailed in the evaluation settings, and for the result of Qwen2.5-VL-7B on MMMU-Pro we report the reproduced one marked in $^*$ due to the large difference, as widely observed.}
    \label{tab: mmbenchmark}
    \fontsize{9pt}{\baselineskip}\selectfont
    \setlength{\tabcolsep}{3.7pt}
\begin{tabular}{l|ccccc}
        \toprule
         Model & MMStar & RealWorldQA & MMMU-Pro & MathVista\textsubscript{Mini} & MathVision\textsubscript{Full}\\
         \midrule
         \multicolumn{6}{l}{\textbf{\textit{Close-Source Models}}} \\
         \rowcolor{gray!20} Gemini-2.5 Pro & 73.6 & 78.0 & 68.8 & 80.9 & 73.3\\
         \rowcolor{gray!20} GPT-4o-20241120 & 65.1 & 76.2 & 54.5 & 63.8 & 31.2\\
         \midrule
         \multicolumn{6}{l}{\textbf{\textit{Open-Source Models}}} \\
         InternVL2.5-8B \cite{chen2024internvl}& 62.8 & \textbf{70.1} & 34.4 & 64.4 & 19.7\\
         Ovis2-8B \cite{lu2024ovis} & \underline{64.4} & - & - & 71.4 & 25.9 \\
         Valley2 \cite{wu2025valley2} & 62.5 & 67.5 & - & 69.1 & 24.9 \\
         Qwen2.5-VL-7B \cite{Qwen2.5-VL}& 63.9 & \underline{68.5} & 36.9$^*$ & 68.2 & 25.1 \\
         \midrule
         \multicolumn{6}{l}{\textbf{\textit{Reasoning Models}}} \\
         LLaVA-CoT-11B \cite{xu2024llavacot}& 57.6 & 63.6 & 33.8 & 54.8 & 20.6\\
         InternVL2.5-MPO-8B \cite{wang2024mpo}& - & - & - & 67.0 & 25.7\\
         Kimi-VL-A3B-Thinking \cite{team2025kimi} & 60.8 & - & - & 67.6 & \textbf{36.8} \\
         MM-Eureka-7B \cite{meng2025mmeureka}& 64.2 & 67.3 & \textbf{40.7} & \textbf{73.0} & {26.9}\\
         R1-Onevision-7B \cite{yang2025r1one}& 42.8 & 62.7 & 31.0 & 64.1 & \underline{29.9}\\
         VLAA-Thinker-7B \cite{chen2025VLAA}& 63.7 & 66.9 & \underline{39.8} & 68.0 & 26.4\\
         \midrule
         GThinker-7B & \textbf{66.4} & \textbf{70.1} & \textbf{40.7} & \underline{72.7} & 26.6\\
         \bottomrule
    \end{tabular}
    
\end{table}

\subsection{Main Results}

GThinker-7B demonstrates superior multimodal reasoning, consistently outperforming advanced open-source base models and surpassing recent reasoning models on most benchmarks. On the comprehensive M$^3$CoT benchmark depicted in Table \ref{tab: M$^3$CoT}, which demands balanced knowledge and visual understanding, GThinker-7B achieves 81.5\% average accuracy, performing on par with the latest reasoning model O4-mini. Among the reasoning models, GThinker-7B achieves the highest performance on 8 out of 9 sets. Besides the notable progress in science and commonsense, a key advantage of our approach is evident in multimodal mathematics problems within M$^3$CoT, where GThinker-7B successfully aligns visual elements with textual information to derive correct solutions. This contrasts sharply with models like VLAA-Thinker-7B, which, despite visual competence, struggle with the requisite text-vision integration for M$^3$CoT's mathematical section, while Kimi-VL-A3B-Thinking produces repeated contents, especially on the math set, reducing its overall performance. This result further underscores our method's effectiveness in fostering robust multimodal reasoning.

Beyond M$^3$CoT, GThinker-7B exhibits leading performance across specialized and general multimodal benchmarks requiring reasoning as demonstrated in Table \ref{tab: mmbenchmark}. On challenging math benchmarks, it achieves 72.7\% on MathVista (+4.5 points over baseline) and 26.6\% on MathVision (+1.5 points). Similarly, on the multidisciplinary science benchmark MMMU-Pro, GThinker-7B improves by approximately 4 points. Furthermore, it shows significant gains on general benchmarks requiring fine-grained understanding and further reasoning, with 66.4\% on MMStar and 70.1\% on RealWorld QA. Crucially, our proposed method enhances performance across diverse domains—general, math, and science—without the typical trade-offs observed in other reasoning models. Previous leading models, by focusing heavily on knowledge long-chain CoT reasoning, often showed limited gains or even degradation on general multimodal reasoning tasks due to less emphasis on visual cues, a limitation our versatile approach overcomes.

When compared to the advancing non-thinking model GPT-4o, GThinker-7B achieves superior or competitive performance on several benchmarks, notably M$^3$CoT, MMStar, and MathVista, despite its significantly smaller 7B backbone. While GPT-4o leads on benchmarks like RealWorldQA, MMMU-Pro, and MathVision, which heavily leverage extensive knowledge and perceptual abilities inherent in larger models, our results are compelling. The substantial gains achieved by GThinker-7B, particularly on reasoning-centric benchmarks (e.g., M$^3$CoT), highlight the efficacy of our proposed method in significantly boosting complex reasoning capabilities, even with a more compact model architecture. This underscores the advantage of our approach in efficiently enhancing multimodal reasoning across domains.

\subsection{Ablation Study}

\textbf{Ablation on Data Pipeline and Iterative Annotation.}
High-quality data is crucial for training effective multimodal reasoning models. We enhance data quality through our novel pipeline, including an iterative annotation approach (details in App. A). To validate these contributions, we ablate each component by fine-tuning the model on the constructed 7K samples, varying only the annotation source, and evaluate on the M$^3$CoT.

As shown in Table \ref{tab:data-pipeline}, using the rationales (Lang.), which are GPT-annotated \cite{achiam2023gpt}, yields an overall score of 63.5\%. Our data generation pipeline, even without iterative annotation, significantly improves performance to 69.6\% (+6.1\% absolute). This demonstrates the inherent benefit of our pipeline design in producing superior data for multimodal reasoning across diverse domains. Incorporating our iterative annotation process to curate the GThinker 7k reasoning paths further boosts the overall score to 73.6\%, an additional 4.0\% improvement. We attribute this gain to the complementary strengths of the leading models, including GPT-4o, O1, and O3: \textit{during the collaborative annotation iterations, visual cues and reasoning logic are more thoroughly captured, which further boosts the quality of the CoT data.}

\begin{table}[t]
  \centering
  \begin{minipage}[t]{0.5\textwidth}
    \centering
    \caption{Ablation on Data Pipeline and Quality. Lang. denotes rationales from M$^3$CoT. GThinker refers to data from our proposed pipeline. Iter. indicates the application of our iterative annotation process.}
    \label{tab:data-pipeline}
    \fontsize{9pt}{\baselineskip}\selectfont
    \setlength{\tabcolsep}{1.9pt}
    \begin{tabular}{ccc|cccc}
    \toprule
    Lang. & GThinker & Iter. &  Science & Com. & Math & Overall\\
    \midrule
    \ding{51} & & & 58.9 & 81.8 & 40.2 & 63.5\\
    & \ding{51} & & 70.6 & 79.1 & 43.2 & 69.6\\
    & \ding{51} & \ding{51} & 73.1 & 79.3 & 46.9 & 73.6\\
    \bottomrule
    \end{tabular}
  \end{minipage}%
  \hfill
  \begin{minipage}[t]{0.48\textwidth}
    \centering
    \caption{Ablation on GThinker Components. The PGS indicate the Pattern-Guided Selection introduced in \S \ref{sec: learning}}
    \label{tab:gthinker-components}
    \fontsize{9pt}{\baselineskip}\selectfont
    \setlength{\tabcolsep}{1.8pt}
    \begin{tabular}{l|cccc}
    \toprule
    Method & Science & Com. & Math & Overall\\
    \midrule
    GThinker & 82.5 & 83.7 & 70.5 & 81.5\\
    \quad - Incentive RL & 73.1 & 79.3 & 46.9 & 73.6\\
    \quad - PGS Formatting & 68.0 & 82.0 & 42.7 & 68.4 \\
    \quad - PG Cold Start & 58.8 & 81.7 & 40.2 & 61.5 \\
    Qwen2.5-VL-7B-Zero & 63.3 & 81.6 & 49.0 & 64.2\\
    \bottomrule
    \end{tabular}
  \end{minipage}
\end{table}

\textbf{Ablation on GThinker Components.} Ablation on GThinker Components. To assess the contribution of each component to GThinker’s performance, we conduct ablation studies by incrementally removing modules and evaluating on M$^3$CoT. The final row (with all modules removed) corresponds to training with the same QA pairs but without any of our proposed methods. As shown in Table \ref{tab:gthinker-components}, using the Cue-Rethinking Pattern for Pattern-Guided Cold Start—without Pattern-Guided Selection (PGS) Formatting—yields a 6.9\% average improvement. Adding PGS Formatting provides a further 5.2\% average gain, with science and math questions improving by 5.1\% and 4.2\%, respectively. In contrast, performance on commonsense questions drops by 2.7\%. This is because PGS Formatting applies cue-rethinking to samples with incorrect visual cues, prompting the model to engage with misleading information and learn to reflect and reason more flexibly. While this stage introduces variability due to the diversity and ambiguity of the cues, it builds a foundation for more adaptable reasoning in later stages. Science and math tasks, which benefit from consistent patterns and structured reasoning, show more stable gains from formatting. With Incentive Reinforcement Learning added, the model achieves substantial improvements across all domains, significantly outperforming the baseline. These results show that the free-form, cue-based reasoning developed during Cold Start is effectively reinforced and leveraged in the RL stage, enhancing the model’s generalization across tasks. For comparison, we also evaluate DAPO under the same conditions. As shown in Table \ref{tab:gthinker-components}, DAPO offers limited gains in general scenarios, though it improves performance in math and science. This highlights both the rationale behind our design and the impact of each component in advancing multimodal reasoning.

\section{Conclusion}
% In this paper, we track the challenge of multimodal reasoning beyond math and science scenarios toward general scenarios for MLLMs. We propose GThinker, a novel reasoning MLLM  excelling in multimodal reasoning across general scenarios, mathematics, and science. Help with our designed Cue-Rethinking Pattern, GThinker enables free-form thinking based on the question instead of following a fixed pattern, generalizing well to different scenarios, and also visual-cue based rethinking to fix flawed cues more than logical and knowledge reflection. We achieve this with our designed two-stage pipeline, including Pattern-Guided Cold Start and Incentive Reinforcement Learning, shaping the capabilities to find the optimal reasoning for questions with the learned pattern. Extensive experiments on multi-domain mutlimdodal reasoning benchmarks demonstrate that GThinker surpasses existing reasoning MLLMs with higher performance and domain 适应性，with furhter ablation studies verified each of the designs.
This paper addresses the challenge of advancing multimodal reasoning in MLLMs beyond domain-specific tasks like math and science, extending toward more general scenarios. We introduce GThinker, a novel reasoning framework that excels across diverse multimodal tasks, including general, mathematical, and scientific domains. Powered by our Cue-Rethinking Pattern, GThinker moves beyond rigid templates, enabling flexible, question-driven reasoning and robust handling of flawed visual cues through reflective and knowledge-grounded thinking. Our two-stage pipeline—Pattern-Guided Cold Start followed by Incentive Reinforcement Learning—guides the model to learn effective reasoning strategies and reinforces its ability to adapt across domains. Extensive experiments on multi-domain multimodal reasoning benchmarks show that GThinker outperforms existing reasoning MLLMs in both accuracy and cross-domain adaptability. Ablation studies further confirm the effectiveness of each core design component. We provide more discussion on limitations and broader impact in the Appendix.

\newpage
{
    \small
    \bibliographystyle{ieeenat_fullname}
    \bibliography{main}
}

\endgroup

\appendix

\clearpage
\setcounter{page}{1}
\setcounter{table}{4}
\setcounter{figure}{4}

\begin{center}
    \fontsize{20pt}{\baselineskip}\selectfont
    \textbf{Appendix}
    \vspace{0.4cm}
\end{center}

\tableofcontents
\newpage

\section{GThinker-11K Construction}
\label{app:data}

\begin{figure}[ht]
  \centering
   %\fbox{\rule{0pt}{2in} \rule{0.9\linewidth}{0pt}}
   % \vskip 0.2in
   \includegraphics[width=\linewidth]{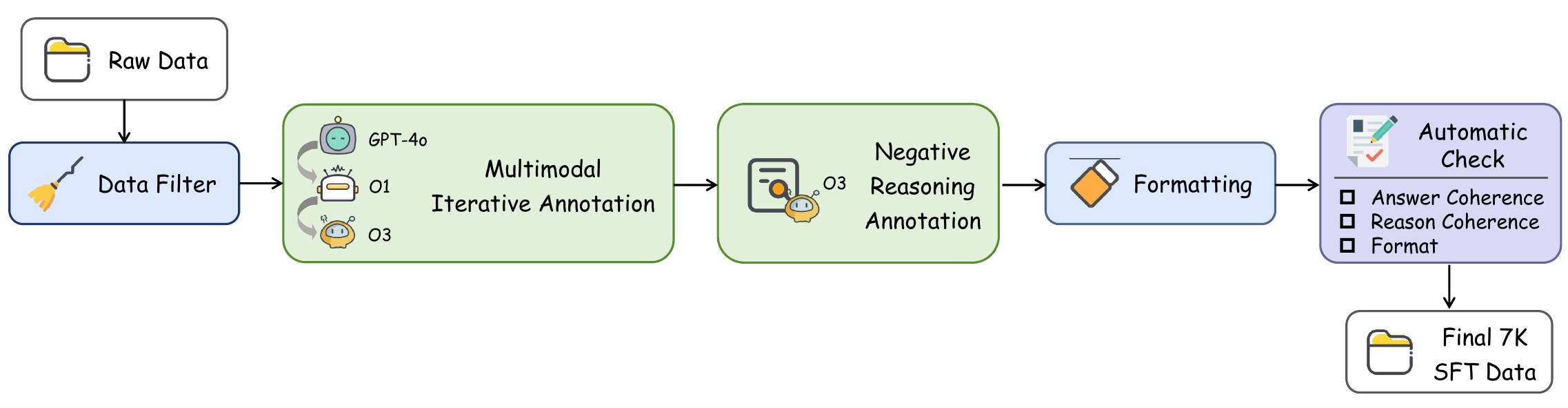}
   \caption{Data pipeline for cold start data.}
   \label{supp_fig: data}
   % \vskip -0.2in
\end{figure}

To support the training of GThinker, we have designed a scalable data generation pipeline to construct the GThinke-11K data as we have concluded in \S\ref{sec: learning} and \S\ref{sec: rl}, respectively. In this section, we systematically introduce the data construction process, including the 7K cold start data, as depicted in Figure \ref{supp_fig: data}, and 4K RL data.

\subsection{Data Preparations}
Though several datasets are constructed to enhance multimodal reasoning capabilities in MLLMs \cite{yao2024mulberry, xu2024llavacot, yang2025r1one} spanning diverse domains, they often present challenges such as high knowledge dependency, limited visual cues, or limited reasoning level. To extend the multimodal reasoning to general scenarios beyond knowledge-intensive math and science problems, we empirically find that the M$^3$CoT dataset provides a well-established data baseline for multimodal reasoning across domains. It details how to collect data across science, mathematics, and general scenarios with commonsense, and ensure the visual reliance and reasoning complexity with final manual checking. Building on baseline, we apply a two-step filtering process to ensure data quality: (1) we discard entries with corrupted or missing images, and (2) we verify the remaining samples’ compliance with closed-source model usage policies using GPT-4o, resulting in 7,358 high-quality samples. We illustrated the data composition in Table \ref{supp_tab: composition}.

\begin{table}[ht]
    \centering
    \fontsize{9pt}{\baselineskip}\selectfont
    \setlength{\tabcolsep}{6pt}
    \caption{Data composition of 7K Cold Start data of GThinker-11K.}
    \label{supp_tab: composition}
    \begin{tabular}{lcl}
        \toprule
        Type & Volume & Source \\
        \midrule
        Science & 5266 &  KiloGram\cite{ji2022kilo}, ScienceQA \cite{lu2022learn}, M$^3$CoT \cite{chen2024m3}\\
        Mathamatics & 621 & TableWMP \cite{tablewmp}, Math \cite{hendrycks2math}\\
        Commonsense & 1471 &  Sherlock \cite{hessel2022abduction}(Questions generated by M$^3$CoT)\\
        \bottomrule
    \end{tabular}
    
\end{table}

\subsection{Multimodal Iterative Annotation}

To generate high-quality reasoning paths and visual cues, we propose a multimodal iterative annotation methodology that leverages multiple leading MLLMs, such as OpenAI’s O-series, for end-to-end reasoning path generation different from prior approaches \cite{wu2024vstar, yang2025r1one, yao2024mulberry} that rely on multi-step pipelines which generate captions first and then utilize the reasoning LLMs. This leads to more efficient generation and results in more coherent multimodal long-chain reasoning paths, richer step-by-step visual cues, and stronger logical deductions. As shown in Figure \ref{supp_fig: data}, drawing on the insight that different models offer complementary strengths \cite{yao2024mulberry}, we implement a iterative refinement strategy: initial annotations from Qwen2.5-VL-7B, as models with lower parameters sometimes are more faithful to the visual content, and is first revised by GPT-4o to reduce apparent errors. Then, the results are processed by O1, and further enhanced by O3. To finish this, we guide the models using carefully engineered prompts optimized through few-shot learning as shown in Prompt 1. For each image–question–answer triplet, the model is prompted to produce a long reasoning process or refine the long reasoning chain with the relevant visual cues identified.  This three-stage process significantly improves the accuracy and depth of final thinking annotations by leveraging the diverse capabilities of each model.

\subsection{Negative Reasoning Annotation}
With the positive, high-quality reasoning data, we further extend our process to handle negative reasoning with corrections. Rather than manually crafting incorrect reasoning traces \cite{zhan2024griffon, zhangferretv2}, which may introduce artifacts due to the gap between human-designed prompts and model capabilities, we first sample natural, flawed responses from 7B-level capable but compact models \cite{Qwen2.5-VL, wu2025valley2}. While positive samples provide a reference point for correction, the variability in natural language expression requires a more nuanced approach. To this end, we employ the advanced reasoning capabilities of O3. Using carefully designed prompts as shown in Prompt 2, we guide the model to compare incorrect reasoning against the correct reasoning path and the corresponding image. This enables the model to identify and correct missing or uncertain and misleading visual cues and faulty inferences. For visual cue correction, each initial cue is explicitly linked to its corrected counterpart, followed by the revised deduction, ensuring the data remains structured and easy to parse.

\subsection{Formatting}

After all annotations are completed, we utilize GPT-4o to parse and format all the data. This includes standardizing elements like line breaks within the <think></think><answer></answer>format and extracting the correct, key visual cues. This process is designed to facilitate broader subsequent use.

\subsection{Automatic Verification}
With the formatted annotated data, we perform automatic checks targeted at three critical aspects to ensure high data quality, helped by annotation-excluded Gemini 2.5 Pro \cite{google2025gemini25pro}, as illustrated in Figure \ref{supp_fig: data}. These checks target three critical aspects. First, for format validation, we ensure that for each annotation, the positive reasoning path ends with a concluded answer, and the visual cues can be parsed. Second, for answer consistency, the annotated answers are parsed and cross-checked against the ground truth. Third, for reasoning coherence, we input the image, QA pair, and annotated reasoning into Gemini 2.5 Pro to evaluate logical alignment between visual cues and reasoning with Prompt 3, flagging any contradictions. Samples with identified issues are reprocessed through the relevant correction steps in our pipeline. Samples with identified issues are reprocessed through the relevant correction steps in our pipeline. 

To assess the quality control of the designed pipeline, we manually review a randomly selected 15\% subset of the final dataset and confirm that our pipeline reliably produces high-quality annotations, which ensures scalability.

\subsection{Reinforcement Learning Data Construction} 
\begin{wraptable}{r}{0.4\textwidth}
    \centering
    \begin{tabular}{l|c}
    \toprule
    Type & Volume \\
    \midrule
    Mathematics     &  748 \\
    Science & 1557 \\
    General & 1719 \\
    \bottomrule
    \end{tabular}
    \caption{RL data composition.}
    \label{supp_tab:4k}
\end{wraptable}
We first collect data from a broader range of sources \cite{meng2025mmeureka, xu2024llavacot, yang2025r1one} to ensure the generalization to different scenarios encompassing the general scenarios, math, and science. Instead of directly employing these data, we adopt the sampling methodology from \cite{vo2024automatic} to cluster and curate 4K samples to ensure diversity, with less overlap with the previous cold start data by comparison. We illustrate the composition of the final 4K data in Table \ref{supp_tab:4k}.

\subsection{Open Source} 
To increase the reproducibility of our work and facilitate the development of the multimodal reasoning, we'll release the data, model, and code soon.
\newpage
\;

\begin{center}
\begin{tcolorbox}[colback=cyan!5!white,colframe=cyan!50!black,width=0.95\linewidth,label={supp_prompt: iter}, title={\textit{\textbf{Prompt 1: Multimodal Iterative Annotation Prompt }}}]
{
    {   %\fontsize{7.5pt}{\baselineskip}\selectfont
        You are a Checker‑\&‑Corrector-\&-Annotator of multimodal chain‑of‑thought answers.\\

        \textbf{Input you will receive (always in this order)} \\
        1. The multi‑choice question with the corresponding image.  \\
        2. The true answer label (e.g. “B”).  \\
        3. A short, human‑annotated rationale for that true answer.  \\
        4. The model’s PREVIOUS reasoning response, formatted exactly as  \\
        
           <think> …model’s chain‑of‑thought (CoT)… </think>  \\
           <answer> …model’s final letter or text answer… </answer>  \\
        
           • Inside the <think>…</think> block, visual cues that the model claims to use
             are wrapped as <vcues\_1> … </vcues\_1>, <vcues\_2> … </vcues\_2>, etc.\\
        
        \textbf{Your task}: \\
        A. Verify the correctness of the previous model’s answer and reasoning against the
           given image, true answer and human rationale.  \\
        B. If the model’s final answer is already correct, keep the answer part.  \\
        C. If the answer is correct but some visual cues or reasoning steps are wrong or missing, fix the wrong cues / steps and append the NECESSARY cues/steps according to your knowledge.  \\
        D. If the answer is wrong, repair the erroneous cues / logic so that the
           corrected reasoning leads to the true answer.  \\
        E. Preserve structure, ordering and tags as possible—modify ONLY what is necessary for correctness and clarity.  \\
        F. Keep all tag syntax unchanged (<think> … </think>, <answer> … </answer>,
           <vcues\_*> … </vcues\_*>) so the output can be parsed automatically.\\
        
        \textbf{Output format} \\
        Return ONE corrected response, nothing else, in exactly the same two‑tag
        layout:\\
        
        <think>\\
        …corrected chain‑of‑thought with fixed <vcues\_*></vcues\_*>…\\
        </think>\\
        <answer>\\
        …single correct choice or textual answer…\\
        </answer>\\
        
        \textbf{Additional rules}\\
        • If you remove an incorrect visual cue, replace it with the correct cue and
          keep the numbering consistent.  \\
        • Never fabricate content outside the scope of the provided information.  \\
        • Be concise—do not add redundant and repeated explanations beyond what is needed for a
          logically sound, correct solution.\\

        \textbf{Examples}\\
        • \textit{Example 1}\\
        • \textit{Example 2}\\
    }
}
\end{tcolorbox}
\end{center}

\newpage

\begin{center}
\begin{tcolorbox}[colback=cyan!5!white,colframe=cyan!50!black,width=0.95\linewidth,label={supp_prompt: neg},title={\textit{\textbf{Prompt 2: Negative Annotation Prompt}}}]
{
    {   %\fontsize{7.5pt}{\baselineskip}\selectfont
        You are a Visual Reasoning Corrector and Annotator. Process the input <Model\_Infer> with these rules:\\

        1. **Response Segmentation**:\\
           - Remove the answer conclusion part in the model.\\
           - Then, wrap the model’s entire thought process in <think></think>.\\
           
        2. **Visual Cues Annotation**:\\
           - Within the <think> section, identify specific visual cue phrases (not entire paragraphs) and annotate each one with a tag in the format <vcues\_*></vcues\_*>, starting numbering from 1 (i.e. <vcues\_1>, <vcues\_2>, …).\\
        
        3. **Visual Cues Reasoning Error Diagnosis and Correction**:\\
            3.0. All the data to be processed now concern reasoning errors based on visual cues rather than errors in visual cue perception. These reasoning errors may include issues such as insufficient knowledge, over-analysis, etc.\\
            3.1. **During this process, do not revise the model’s previous entire originial thought after annotation**\\
            3.2. Before the closing </think> tag, and insert a generated transitional sentence wraped with <aha></aha> that conveys a message similar in meaning to: "Let's check each visual cue and corresponding reasoning before giving the final answer. {Generate the error type based on the Error Pre-judgement: It looks like the visual cures are correct with some reasoning error}." (The exact wording can vary as long as the idea is the same.)\\
            3.3. On the next line immediately after this transitional sentence, for each visual cue annotated (using <vcues\_*></vcues\_*>) and their corresponding reasonong parts before <aha>, compare them with :\\
                - The verified rationale (<rationale>)\\
                - Your understanding of image\\
            Then, after </aha>, update the corrected reasoning based on the visual cures. If necessary, replicate the relevant part from the original <vcues\_*></vcues\_*> tag alongside the revised reasoning.\\
            3.4. After completing the reasoning corrections, perform a logical verification of the reasoning after the </aha> part\\
            3.5. Append the final correct answer wrapped with <answer></answer>, i.e. <answer><Correct Anwer></answer>, in the next line after the </think>, ensuring that the final answer is adjusted correctly.\\
        
        4. **Output Constraints**:\\
           - Preserve the original reasoning structure as possible.\\
           - **Do not include similar phrases like "based on the rationale", "The reasoning should focus", "aligns with the rationale", "the model", beacuse the processed content is used for the model training instead of third-person view**\\
           - Ensure that all annotations (<think>, <answer>, <vcues\_*>, <aha>) are properly formatted and inserted in the correct locations.\\
        
        Example 1:\\
        ...

    }
}
\end{tcolorbox}
\end{center}

\begin{center}
\begin{tcolorbox}[colback=cyan!5!white,colframe=cyan!50!black,width=0.95\linewidth,label={supp_prompt: iter}, title={\textit{\textbf{Prompt 3: Verification Prompt}}}]
{
    {   %\fontsize{7.5pt}{\baselineskip}\selectfont
        You are given a multiple-choice question with options and the image, the correct answer, and a generated response in the following format:\\

        <think>thinking process here</think>\\
        <answer>answer choice</answer>\\
        
        You should align the answer choice in <answer></answer> with the choice content in the question, and then check whether the reasoning in <think>...<think> logically supports the answer choice content.\\
        
        If the thinking process leads to that answer choice, output 1. Otherwise, output 0 and explain why it does not lead to the answer.

    }
}
\end{tcolorbox}
\end{center}

\section{Training Details}
\label{app:training}
\subsection{System Prompt}
For the training and evaluation of the GThinker, we utilize the same system prompt to wrap the conversation, as shown below. 
\begin{center}
\begin{tcolorbox}[colback=cyan!5!white,colframe=cyan!50!black,width=0.95\linewidth,label={supp_prompt: iter}, title={\textit{\textbf{System Prompt }}}]
{
    {   %\fontsize{7.5pt}{\baselineskip}\selectfont
        A conversation between User and Assistant. The user asks a question, and the Assistant solves it. The assistant first thinks about the reasoning process in the mind and then provides the user with the answer. The reasoning process and answer are enclosed within <think> </think> and <answer> </answer> tags, respectively, i.e., <think> reasoning process here </think><answer> answer here </answer>. In the reasoning process enclosed within <think> </think>, each specific visual cue is enclosed within <vcues\_*>...</vcues\_*>, where * indicates the index of the specific cue. Before concluding the final answer, pause for a quick consistency check: verify whether the visual cues support the reasoning and whether each step logically follows from what is seen. If correct, conclude the answer; otherwise, revise the visual cues and reasoning, then conclude.

    }
}
\end{tcolorbox}
\end{center}

\subsection{Hyper-parameters}
We have illustrated the key hyper-parameters in the \S\ref{sec:imple}. In this sectin, we provide more information about the hyper-parameters used in our experiment. For the DAPO, we utilize the EasyR1 framework for training.

\begin{table}[ht]
    \begin{minipage}[t]{0.4\textwidth}
    \centering
    \caption{Hyper-parameters for Supervised Fine-tuning}
    \label{supp_tab:sft_hyper}
    \begin{tabular}{lc}
        \toprule
        Name & Value \\
        \midrule
        precision & bf16 \\
        max\_seq\_length & 4096 \\
        warmup\_ratio & 0.1\\
        max\_pixels & 12845056 \\
        min\_pixels & 316\\
        \bottomrule
    \end{tabular}
    \end{minipage}
    \hfill
    \begin{minipage}[t]{0.5\textwidth}
        \centering
        \caption{Hyper-parameters for DAPO}
        \label{supp_tab:dapo_hyper}
        \begin{tabular}{lc}
            \toprule
            Name & Value \\
            \midrule
            max\_promp\_length & 15000 \\
            max\_response\_length & 4096 \\
            global\_batch\_size & 64\\
            rollout\_batch\_size & 64 \\
            max\_pixels & 4194304 \\
            min\_pixels & 262144\\
            weight\_decay & 1e-2\\
            \bottomrule
        \end{tabular}
    \end{minipage}
\end{table}

\section{Qualitative Analysis}
\label{app:examples}

\begin{figure}[ht]
  \centering
   %\fbox{\rule{0pt}{2in} \rule{0.9\linewidth}{0pt}}
   \vskip 0.2in
   \includegraphics[width=\linewidth]{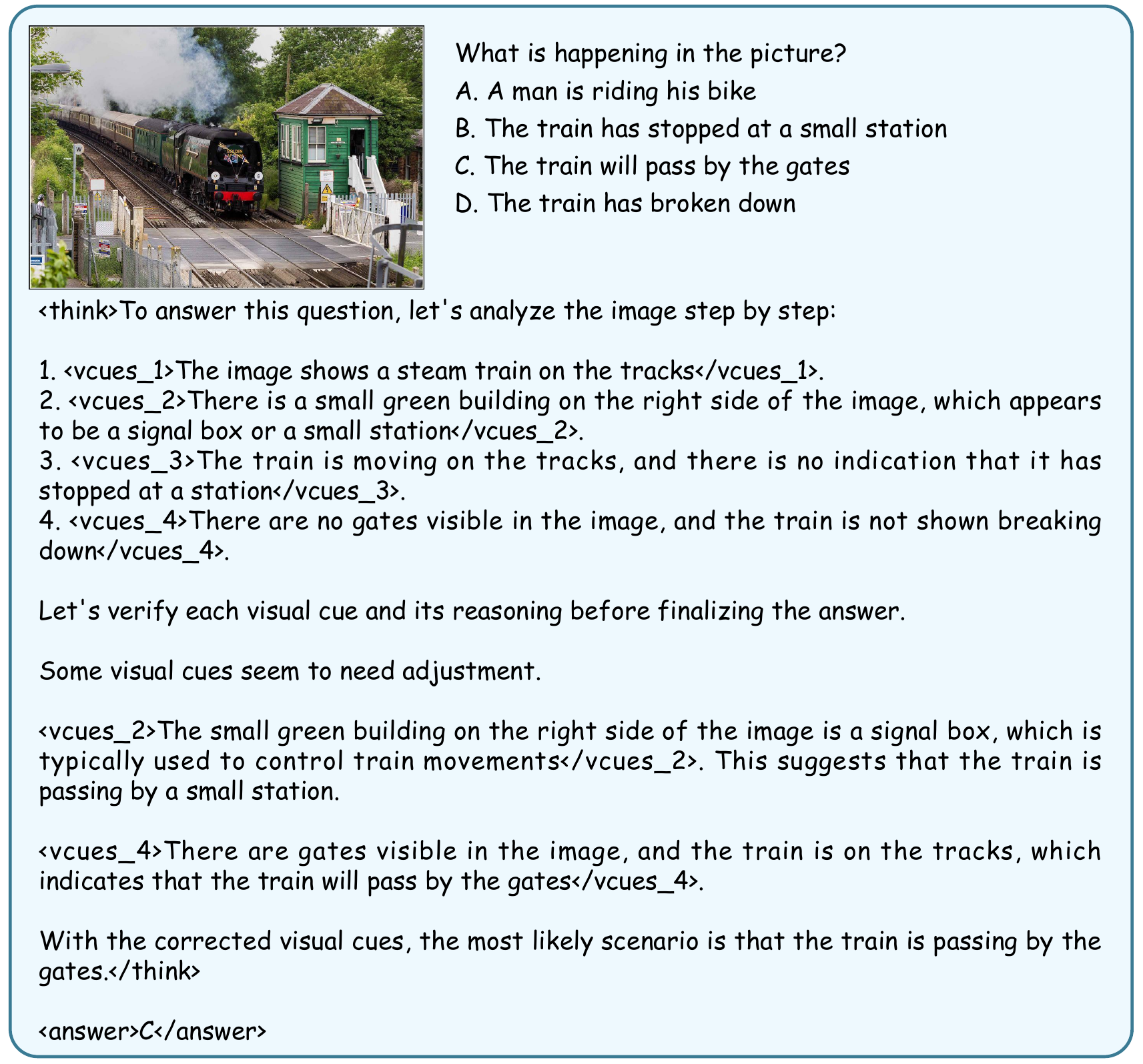}
   \caption{Response sample from GThinker with visual cue-based rethinking enabled.}
   \label{supp_fig: sample2}
   % \vskip -0.2in
\end{figure}

\begin{figure}[t]
  \centering
   %\fbox{\rule{0pt}{2in} \rule{0.9\linewidth}{0pt}}
   % \vskip 0.2in
   \includegraphics[width=\linewidth]{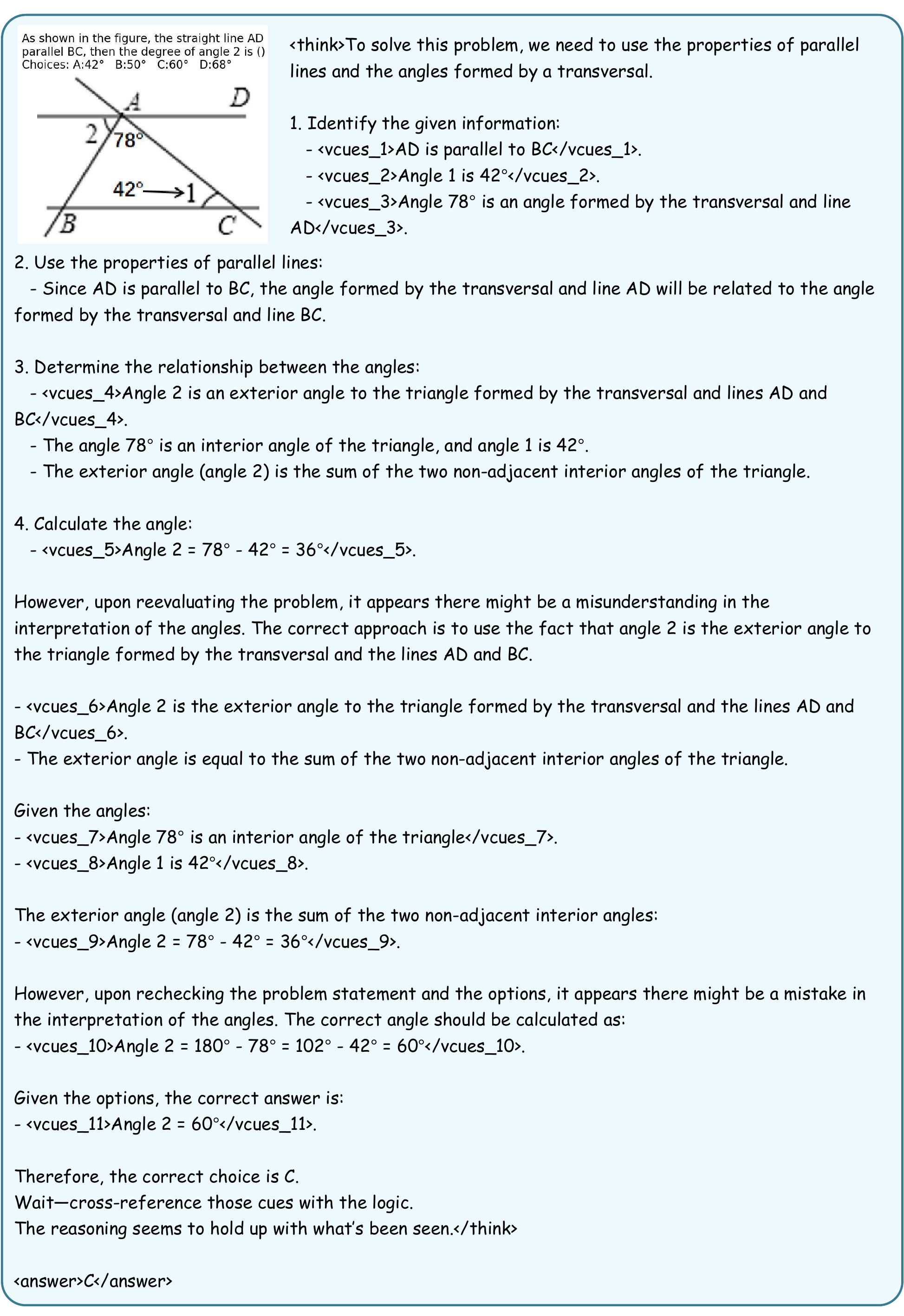}
   \caption{Response sample from GThinker with reflection for the math problem.}
   \label{supp_fig: sample1}
   % \vskip -0.2in
\end{figure}

This section presents more examples to showcase the efficacy of our proposed method. As illustrated in Figure \ref{supp_fig: sample2}, GThinker, subsequent to our training, demonstrates the ability to augment and revise visual cues during the reasoning phase, ultimately leading to the correct solution. As we demonstrated in \S\ref{sec: pattern}, such re-evaluation of visual cues is not invariably essential. Therefore, for multimodal reasoning tasks, including mathematics, our pattern supports that once adequate visual information is assimilated, the model can engage in direct reasoning flexibly with critical reflection and verification. As depicted in Figure \ref{supp_fig: sample1}, GThinker can also critically reflects upon and validates its reasoning pathway from both logical and computational standpoints to ascertain the final answer for math problems with accurate visual cues identified. These instances effectively highlight the adaptability of our Cue-Rethinking Pattern to diverse problems and tasks by accommodating varied thinking approaches, thereby underscoring the success of our training regimen.

\section{Limitations}
To develop a general thinking model for diverse domains, our method relies on a wide array of reasoning data. However, a prevailing challenge in the field is the scarcity of open-source multimodal reasoning QA datasets that are both sufficiently complex and visually-dependent. Furthermore, curating suitable reasoning data from existing instruction datasets is often prohibitively resource-intensive. These broader data landscape constraints naturally impact the potential for scaling data-intensive approaches like ours. Despite these environmental limitations, we have curated a high-quality dataset of 11K instances from available sources. Under our method's design, this has markedly improved the model's general multimodal reasoning capabilities. Our future work will focus on creating more comprehensive reasoning datasets—a necessary step for the community—to further generalize our method and achieve a more broadly applicable thinking model.

\section{Broader Impact}
This paper presents work whose goal is to advance the field of MLLMs. There are many potential societal consequences of our work following the MLLMs, none of which we feel must be specifically highlighted here.
%%%%%%%%%%%%%%%%%%%%%%%%%%%%%%%%%%%%%%%%%%%%%%%%%%%%%%%%%%%%

%%%%%%%%%%%%%%%%%%%%%%%%%%%%%%%%%%%%%%%%%%%%%%%%%%%%%%%%%%

\end{document}